# An Integrated Neighborhood and Scale Information Network for Open-Pit Mine Change Detection in High-Resolution Remote Sensing Images

Zilin Xie [a], Kangning Li [a,*], Jinbao Jiang [a], Jinzhong Yang [b], Xiaojun Qiao [a], Deshuai Yuan [c], Cheng Nie [a]

[a] College of Geoscience and Surveying Engineering, China University of Mining and Technology-Beijing, Beijing 100083, China

[b] China Aero Geophysical Survey & Remote Sensing Center for Nature Resources, China Geological Survey, Beijing 100083, China

[c] National Engineering Laboratory for Satellite Remote Sensing Applications, Aerospace Information Research Institute, Chinese Academy of Sciences, Beijing 100083, China

* Corresponding author: Kangning Li (202213@cumtb.edu.cn)

## Abstract

Open-pit mine change detection (CD) in high-resolution (HR) remote sensing images plays a crucial role in mineral development and environmental protection. Significant progress has been made in this field in recent years, largely due to the advancement of deep learning techniques. However, existing deep-learning-based CD methods encounter challenges in effectively integrating neighborhood and scale information, resulting in suboptimal performance. Therefore, by exploring the influence patterns of neighborhood and scale information, this paper proposes an Integrated Neighborhood and Scale Information Network (INSINet) for open-pit mine CD in HR remote sensing images. Specifically, INSINet introduces 8-neighborhood-image information to acquire a larger receptive field, improving the recognition of center image boundary regions. Drawing on techniques of skip connection, deep supervision, and attention mechanism, the multi-path deep supervised attention (MDSA) module is designed to enhance multi-scale information fusion and change feature extraction. Experimental analysis reveals that incorporating neighborhood and scale information enhances the F1 score of INSINet by 6.40%, with improvements of 3.08% and 3.32% respectively. INSINet outperforms existing methods with an Overall Accuracy of 97.69%, Intersection over Union of 71.26%, and F1 score of 83.22%. INSINet shows significance for open-pit mine CD in HR remote sensing images.

## 1. Introduction

Mineral resources are important for the development of human society and play a significant role in national economic construction. However, the exploitation of mineral resources, especially in open-pit mines, has caused severe damage to the ecological environment (Chen et al., 2018; Wilkins et al., 2020). Therefore, it is necessary to conduct change detection (CD) in open-pit mines to facilitate the sustainable utilization of mineral resources. Remote sensing images enable extensive and high-frequency monitoring of mining areas (Kumar and Gorai, 2023). Currently, CD methods based on remote sensing images have been widely applied in monitoring open-pit mines (Barenblitt et al., 2021; Gao et al., 2021; Nie et al., 2022).

The emergence of high-resolution (HR) remote sensing data has provided strong support for more refined CD. However, compared to general remote sensing images, greater complexity and diversity of high-resolution remote sensing images might result in larger intra-class differences and smaller inter-class differences (Wang et al., 2022; Lv et al., 2022). In addition, mining areas contain a variety of feature elements such as ore, waste rock, bare soil and buildings, making the scene more complicated (Pan et al., 2023). Therefore, the current CD methods for HR remote sensing images of open-pit mines still mainly focus on manual visual interpretation by professionals (Du et al., 2022a). Other traditional methods mainly focus on studying and analyzing the extracted image features such as spectra, texture, and shape, and then performing remote sensing monitoring (Blachowski et al., 2023; Fekete and Cserep, 2021; Hao et al., 2023; Padmanaban et al., 2017; Romary et al., 2015). However, these manually designed features might be insufficient for the expression and differentiation of land features in complex mine environments (Du et al., 2021), and the accuracy of the model could be significantly diminished following alterations in data or study areas (Chen et al., 2023).

In recent years, deep-learning-based techniques have gained increasing attention due to their powerful ability to intelligently extract deep information of images (Jiang et al., 2022; Liu et al., 2023, 2021; Zhang et al., 2022), and have found wide application in the field of open-pit mine CD for HR remote sensing images (Camalan et al., 2022; Du et al., 2022a; Gallwey et al., 2020; Tang et al., 2021). However, most current methods ignore the issues of restricted receptive field and incomplete information of objects, leading to poor recognition of image boundary regions. As shown in Fig.1 (a), the lack of neighborhood information in the red circle might lead to mountains being mistaken for mine areas. This challenge is not confined to open-pit mine CD but also exists in other applications of deep learning in HR remote sensing. This arises from the fact that HR remote sensing images are much larger compared to common natural images in the computer vision domain. For instance, the images in the PASCAL VOC natural image dataset (Everingham et al., 2015) are of size 500 × 375 pixels, whereas those in the WHU remote sensing image dataset (Ji et al., 2019) are of size 15354 × 32507 pixels. When dealing with HR remote sensing images, scholars are compelled to pre-crop the images to an appropriate size and stitch the outputs as the final prediction; otherwise, the computational cost would be prohibitive (Ding et al., 2020; Zhang et al., 2022). This phenomenon is even more serious in open-pit mine detection task. Mine objects are much larger compared to general building objects, making it difficult to obtain a full view of mine objects in cropped images (Xie et al., 2023).

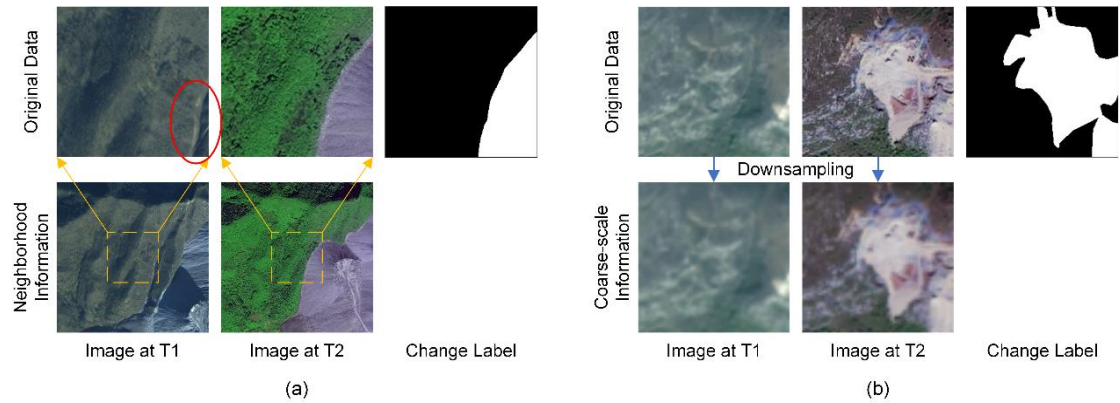

Fig. 1. Influence of neighborhood and scale information. (a) Influence of neighborhood information. (b) Influence of scale information.

Currently scholars are attempting to address this issue. The most common solution is to set a certain image overlap region during cropping (Du et al., 2022b; Liu et al., 2018). However, increasing the overlap rate also increases the time consumption and requires additional post-processing to determine the final result. Another common method is to set up mirror padding or other methods to expand the image, thereby enhancing the recognition of image boundary regions (Ronneberger et al., 2015). However, the boundary regions generated by this method are not realistic and ignore the spatial correlation between the cropped remote sensing images. Besides, some scholars have attempted to address the problem from a model perspective. Zhang et al. (Zhang et al., 2022) tried to directly input a very large-sized remote sensing image (e.g., an image size of 2448 pixels) and constructed a model adapted to large image inputs. Ding et al. (Ding et al., 2020) constructed an embedded auxiliary network designed to learn high-level features from the downsampled large image, thereby aiding in the detection of normal-sized images within that large image. Despite both of the above methods expanding the receptive field through larger images, they still cannot avoid the detection errors at the boundary regions of the large images.

In addition to the aforementioned issue of detection at the image boundary regions, optimization should also be conducted for the potential scale problem between HR remote sensing images and mine areas. As shown in Fig.1 (b), although a coarse scale reduces the clarity of features, it concurrently diminishes noise and enhances the integrity of the mining area. This implies that higher resolution does not necessarily yield better detection results. Various methods have been proposed to address the scale problem in images. For instance, skip connections (Ronneberger et al., 2015; Zhou et al., 2018) can be used to combine shallow features with deeper features, thus comprehensively utilizing features of different scales. Deep supervised strategies (Wang et al., 2015; Zhang et al., 2020) introduce supervised signals in the middle layer of the network, enabling the network to learn more discriminative features in different scale feature layers. Attention mechanisms (Vaswani et al., 2017; Woo et al., 2018) facilitate the model in dynamically adjusting its attention to different regions or scales in the input, thus capturing important information more efficiently. Despite the existence of various methods to tackle scale problems, most of them are designed for common surface features, such as buildings, farmland, forests, etc., with relatively less research on open-pit mine. The influence pattern of scale information in HR remote sensing CD for open-pit mines is not yet clear and requires further exploration.

Based on the analysis of the above problems, this paper optimizes the open-pit mines CD in HR remote sensing images using neighborhood and scale information. Firstly, through exploring and summarizing the neighborhood and scale information, the patterns of information influence in this task are identified to guide subsequent model construction. Secondly, neighborhood information is obtained through the 8-neighborhood images around the image, assisting in multi-scale feature extraction of the center images. In this process, the center and neighborhood images are fed into a Quadruple network for feature extraction, and feature enhancement and fusion are achieved through time fusion (TF) and center-neighborhood fusion (CNF) modules. Subsequently, leveraging techniques of skip connection, deep supervision and attention mechanism, a multi-path deep supervised attention (MDSA) module is designed to aggregate multi-scale features. Finally, an Integrated Neighborhood and Scale Information Network (INSINet) is proposed. INSINet also considers model lightweighting, realizing efficient extraction of change information from HR remote sensing images of open-pit mines.

The contributions of this paper are as follows:

(1) The influence patterns of neighborhood and scale information on open-pit mine CD in HR remote sensing images are explored.

(2) Neighborhood image information is introduced to achieve a larger receptive field and alleviate challenges of poor recognition in the boundary regions of images.

(3) A MDSA module is designed to enhance the fusion of multi-scale information and the extraction of change features.

(4) An INSINet is proposed based on the integration of neighborhood and scale information, achieving optimal performance in the CD of HR remote sensing images of mining areas.

The rest of this paper is organized as follows. Section II provides a detailed description of the proposed INSINet. Section III introduces experiments, including neighborhood and scale information analysis, as well as method comparisons. Section IV further discusses the influence of neighborhood and scale information. Finally, Section V concludes this paper.

## 2. Methods

### 2.1. Overall architecture of INSINet

As shown in the Fig. 2, the INSINet model consists of three parts: neighborhood information acquisition, feature extraction, and multi-scale information fusion. Firstly, in the neighborhood information acquisition stage, the 8-neighborhood images around the image are acquired and combined into a larger image. The larger image is then downsampled to a standard image size, serving as the source of neighborhood information acquisition. In the feature extraction stage, center and neighborhood images from two time points are fed into a Quadruple network consisting of four encoders. Subsequently, the bitemporal features are fused through the TF module, and then the center and neighborhood features are fused through the CNF module to extract features at each scale. Finally, in the multi-scale information fusion stage, a main pathway and multiple branch pathways are included. In the main pathway, preliminary predicted result is obtained from the center feature through a decoder. In the branch pathways, preliminary prediction results are also computed from the features extracted by the CNF at different scales, and then fed into the MDSA module to enhance the main pathway. A multi-scale aggregation (MSA) module aggregates above preliminary results of different paths and scales into a final change result.

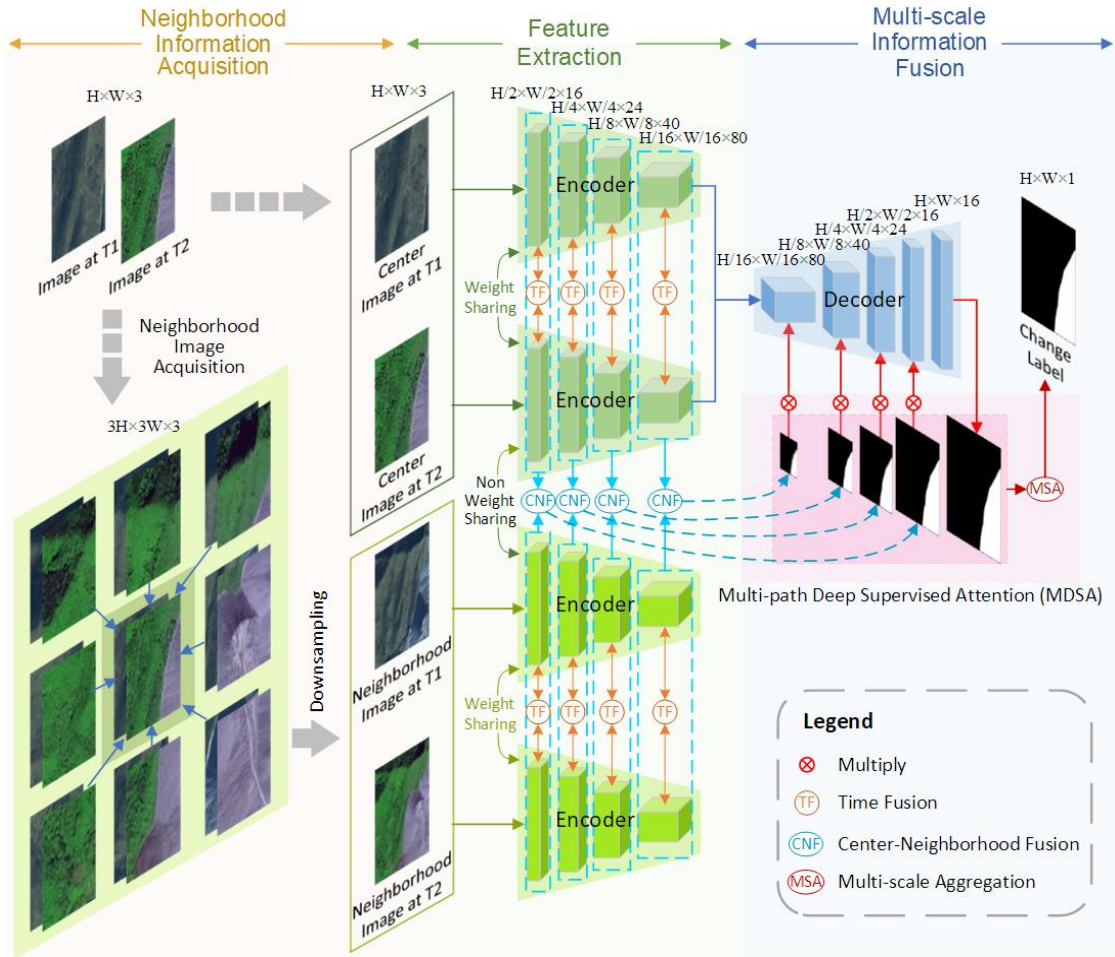

Fig. 2. Overall architecture of INSINet. The INSINet model consists of three parts: neighborhood information acquisition, feature extraction, and multi-scale information fusion.

The INSINet model is also designed with lightweighting in mind, addressed by leveraging MobileNetV3 (Howard et al., 2019). In this paper, the first 11 layers of MobileNetV3-Large, proposed by MobileNetV3, are used as the network backbone, thereby achieving lightweight encoders while ensuring accuracy. In addition, the Bottleneck module from MobileNetV3 is adopted as the primary convolution module to facilitate the lightweight design of other parts of INSINet.

## 2.2. Neighborhood information acquisition

As shown in Fig. 2, the neighborhood information is acquired by retrieving the 8-neighborhood images surrounding the center images. However, this process is not highly efficient in practical applications. It is more cost-effective to pre-crop the neighborhood images corresponding to the center images. Performing the acquisition of neighborhood information as a pre-processing step can save the time of searching neighborhood images.

In addition, the neighborhood images are downsampled before being input into the network. This is because these images are primarily used to extract neighborhood information and do not require high resolution, while the center images are used to extract precise change information. Moreover, in the application of open-pit mine CD in HR remote sensing images, higher resolution does not necessarily lead to better detection results. Appropriately reducing the resolution may actually improve the performance of CD models, as detailed in Section 3.3.2. Therefore, the neighborhood images are downsampled to the standard image size to facilitate the reduction of model parameters and the subsequent matching and fusion of the center and neighborhood information.

## 2.3. Feature extraction

### 2.3.1. Quadruple network

Siamese network, consisting of two identical encoders, is a common structure for CD, used to enhance the analysis of differential features in bitemporal images (Chen et al., 2022). Siamese networks can be categorized into weight-sharing and non-weight-sharing strategies. The weight-sharing strategy can optimize memory and training time while maintaining performance (Varghese et al., 2019). In contrast, the non-weight-sharing strategy enhances the feature extraction of each part but requires more memory and time consumption (Lei et al., 2023).

The Quadruple network proposed in this paper contains four identical encoders. It can be viewed as two pairs of Siamese encoders. Each pair shares weights internally, while there is no weight sharing across the two pairs. This design is related to the structure of the center and neighborhood data. The internal spatial extents and scales of bitemporal data in the center or neighborhood are consistent, enabling the utilization of a weight-sharing strategy for optimizing the network structure. However, the spatial extents and scales between the data of the center and the neighborhood are inconsistent, requiring the use of a non-weight-sharing strategy for individual feature extraction.

### 2.3.2. TF

The structure of the TF module is illustrated in Fig. 3 (a). The TF module is not only used for temporal information fusion between weight-sharing encoders but also for providing feedback to the encoders. The temporal information interaction between encoders has the potential to suppress task-irrelevant interference, thereby motivating encoders to focus more on the areas of truly changes (Feng et al., 2023).

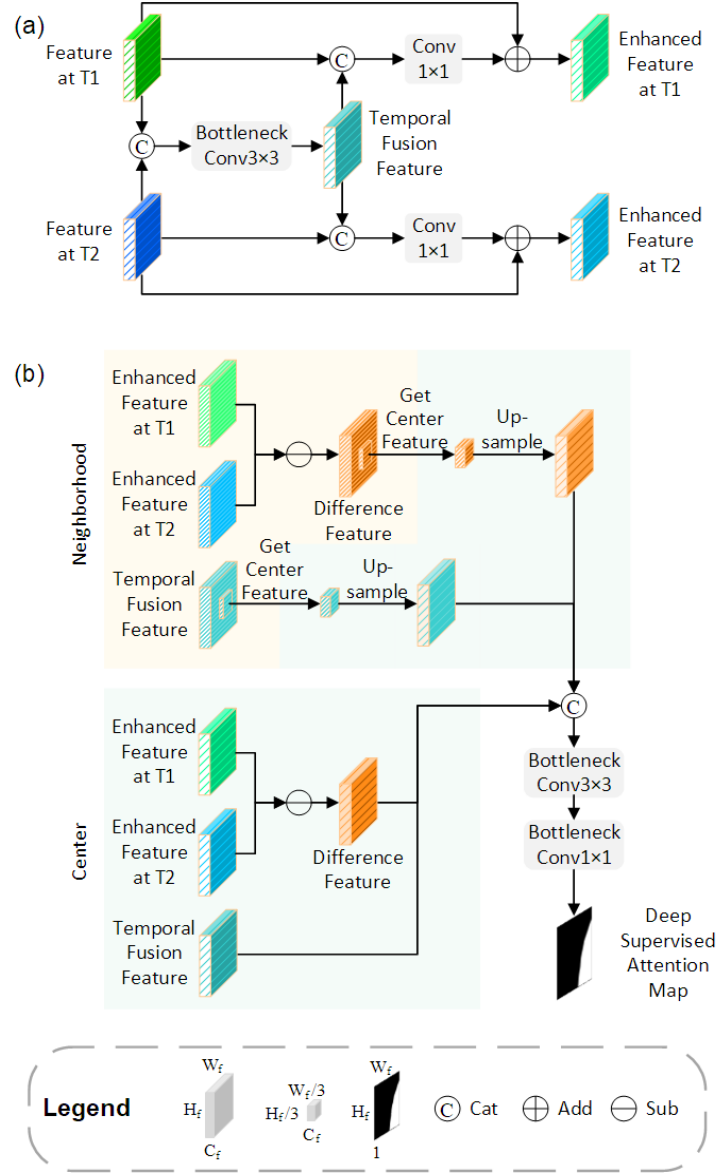

Fig. 3. Feature fusion modules. (a) TF module for interaction of bitemporal information. (b) CNF module for aggregation of center and neighborhood information.

2.3.3. CNF

The structure of the CNF module is illustrated in Fig. 3 (b). The center and neighborhood features, acquired respectively through the TF module, are input together into the CNF module. Considering that the TF features contain common information of the bitemporal features, the difference features of the enhanced bitemporal features are computed. This processing ensures effective feature extraction while reducing the model parameters, as detailed in Section 3.6. It is worth noting that the spatial coverage of the center features and the neighborhood features are inconsistent. To address this, the regions corresponding to the center features are extracted from the neighboring features and subsequently upsampled to restore their original size. This procedure ensures uniform spatial coverage between the center and neighboring features. Finally, the center and neighborhood information are fused and mapped into a deep supervised attention map for subsequent MDSA modules.

### 2.4. Multi-scale information fusion

#### 2.4.1. MDSA

As shown in Fig. 2, the MDSA module draws inspiration from techniques of skip connection, deep supervision and attention mechanism, but differs from them in several aspects. Firstly, the multi-scale features of encoders are transmitted to the corresponding scales of decoder, which serves as the main pathway, through the skip connections. However, instead of being incorporated into the main pathway as conventional approaches (Chen et al., 2023; Du et al., 2022a; Xiang et al., 2022), it is directly calculated to obtain initial predicted results at different branch pathways. These predictions are then aggregated into the final results by the subsequent MSA module. This idea of decision-making based on the integration of the multi-path predictions can be regarded as constructing multiple classifiers for ensemble learning (Ganaie et al., 2022), enhancing flexibility and adaptability of the model to objects of different scales. Secondly, the deep supervision of MDSA also differs from general methods (Deng et al., 2023; Li et al., 2022; Wu et al., 2023) that output various deep features of the main path for calculating loss. The deep features of MDSA are extracted from four branch paths at different scales and serve as attention maps to reinforce the main path. This strategy avoids the redundancy of the supervision effect caused by the method of supervising only the main path. It also strengthens the effect of feature extraction from each branch while ensuring the performance of the deep features of the main path, and passes the discriminative information to the encoders through skip connections. Thirdly, MDSA introduces discriminative information to assist attention maps in focusing the model more on the change areas. Compared to conventional attention mechanisms without discriminative information (Song and Jiang, 2021; Song et al., 2022; D. Wang et al., 2022), MDSA exhibits a stronger capability in guiding change information. In summary, MDSA achieves superior multi-scale information fusion and change information extraction through the organic combination of skip connection, deep supervision and attention mechanism.

### 2.4.2. MSA

The MSA module serves as the deep supervision component within the MDSA module, facilitating the automatic adjustment of the weights between the multi-scale initial predicted results as well as the further optimization of the final results. As illustrated in Fig. 4, the MSA module initially upsamples the multi-scale preliminary predictions to their original sizes. Subsequently, it achieves interaction of multi-scale predictions through 1×1 and 3×3 convolutions, aggregating the final predicted results.

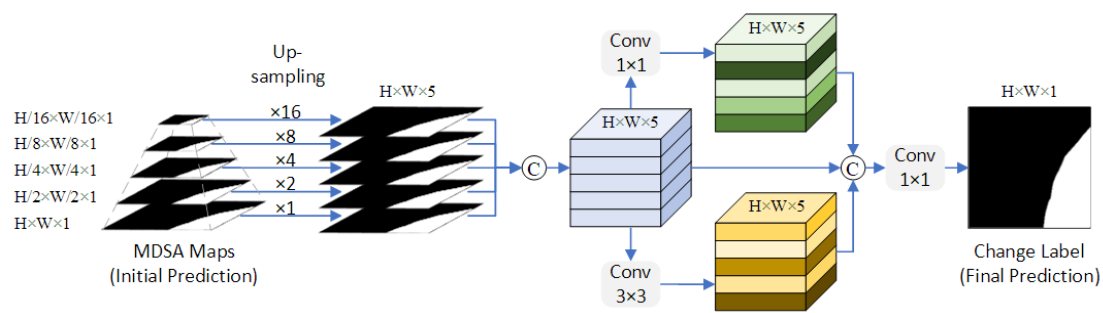

Fig. 4. The MSA module aggregates multi-scale initial predictions for final prediction.

### 2.5. Loss function

Due to the integration of multi-scale deep supervised predictions into a finial prediction by the MSA module, there is no need to use the deep supervised loss function (Zhang et al., 2020). Instead, a general loss function is sufficient. In this paper, minimized cross-entropy loss is utilized to optimize the network parameters. The loss function is defined as follows:

$$L = \frac{1}{H \times W} \sum_{h=1,w=1}^{H,W} l(P_{hw}, Y_{hw}) \quad (1)$$

where $l(P_{hw}, y) = -\log(P_{hwy})$ is the cross-entropy loss, and $Y_{hw}$ denotes the label of the pixel at position $(h, w)$.

## 3. Experiment results

### 3.1. Datasets

#### 3.1.1. Open-pit mine CD dataset

In this study, an open-pit mine CD dataset, as illustrated in Fig. 5, was created using multi-source fused HR remote sensing images and mining rights vector data from partial counties of Liaoning Province, China. These images were acquired during 2018 and 2019, covering the period from June to December each year, and consisted of a multi-source fusion of satellite data, including GF-1, GF-2, GF-6, Beijing-2, and ZY-3, with a spatial resolution ranging from 0.5 to 2 meters. The mining rights vector data, accumulated over the years by the China Geological Survey, served as auxiliary information for discriminating mining areas. After being pre-processed, such as image registration, these images were visually interpreted by experts to draw change maps by relevant standards ("Specification for Investigation and Evaluation of Mine Geological Environment," 2014; "Technical Requirements for Remote Sensing Monitoring of Mineral Resource Development," 2011).

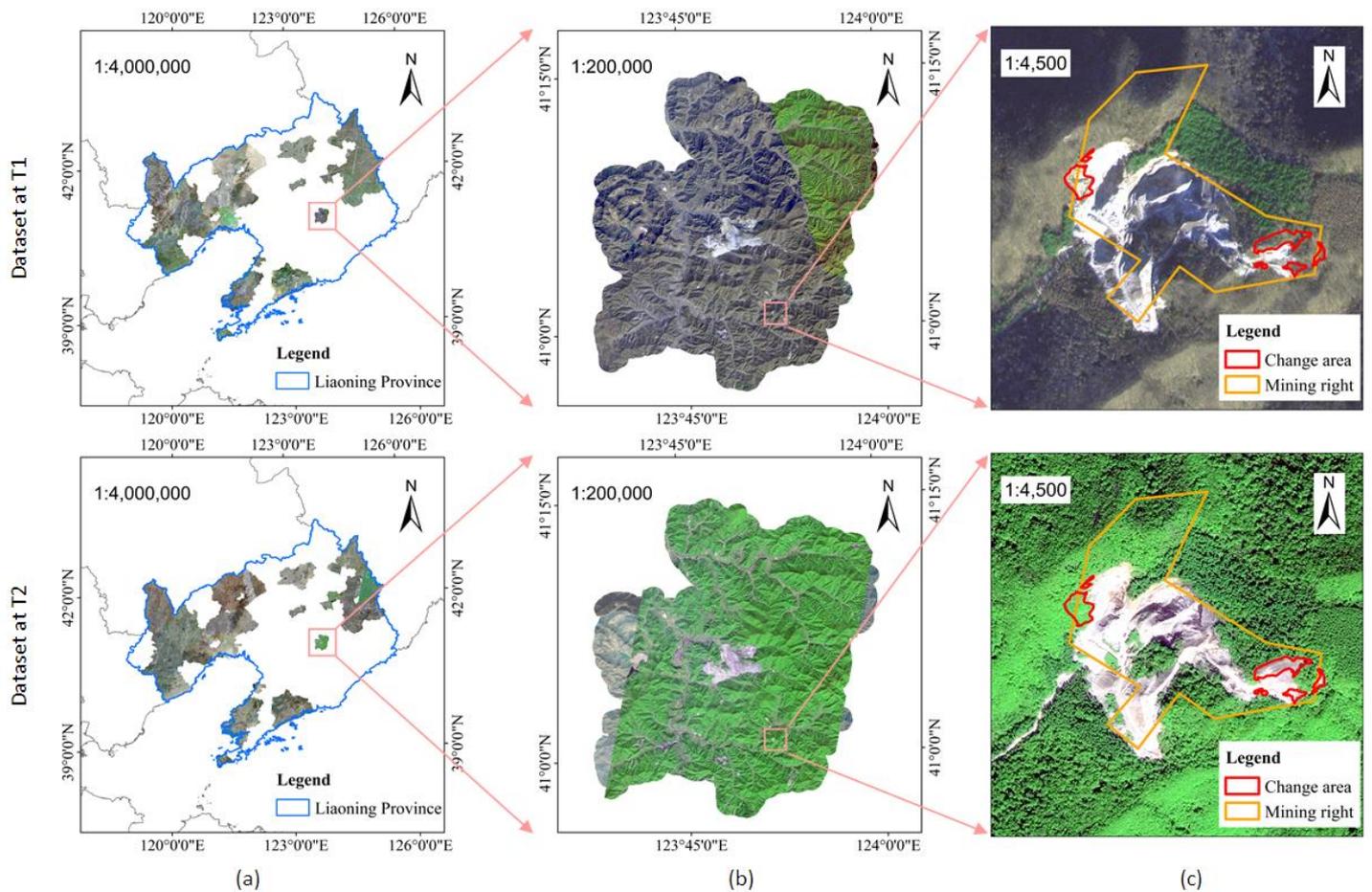

Fig. 5. Illustration of open-pit mine CD dataset with an example of a mining area in Nanfen County, Liaoning Province. (a) Selected counties in Liaoning Province covered by the dataset. (b) Multi-source fused HR remote sensing images of Nanfen County. (c) HR remote sensing images, mining right vectors, and change area vectors of a mining area.

The images were resampled to a resolution of 1 m, and cropped into image patches of 256 × 256, resulting in 2693 sets of open-pit mine CD dataset. The labels were binary maps, indicating whether have changes occurred. The CD dataset was randomly divided into training, validation and test sets in a ratio of 6:2:2. To mitigate the impact of data scarcity on the model, data augmentation techniques were employed, including image flipping, image rotation, color transformation, and increased overlap rate (Xie et al., 2024). These techniques resulted in an increase of the training set to over 1000 times its original size and were implemented automatically through code.

### 3.1.2. Neighborhood information analysis dataset

To analyze the influence of neighborhood information on the effectiveness of open-pit mine CD in HR remote sensing images, a neighborhood information analysis dataset was created by modifying the test set of open-pit mine CD dataset. As shown in Fig. 6, each image was roughly divided into four areas according to the amount of neighborhood information: the outer ring, middle ring, inner ring and core area. The width of each ring equaled 1/8 of the image side length. However, land features vary across different locations of the image. To ensure the consistency of the detected features, a 1/8H × 1/8W pixel region in the image was randomly selected as the detection region. Subsequently, images of standard size were cropped from the large-extent remote sensing image as the new test data, corresponding to the detection regions with the same features but different locations. Moreover, to ensure the stability of the neighborhood information analysis, the average CD accuracy of the four corner regions (four 1/8H × 1/8W pixel regions) in each neighborhood area was selected as the final accuracy. Therefore, it was necessary to generate 16 new test sets and assemble them into 4 groups of test sets to assess the CD effectiveness in four different neighborhood information scenarios, thereby analyzing the influence pattern of neighborhood information.

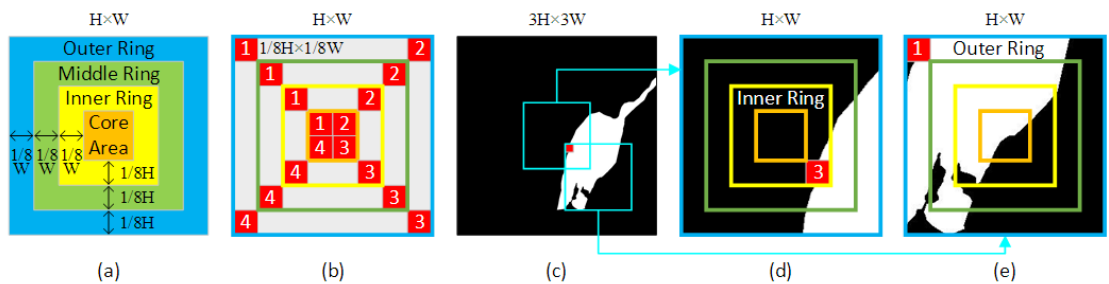

Fig. 6. Illustration of neighborhood information analysis dataset. (a) The image is roughly divided into four areas according to the amount of neighborhood information. (b) The average CD accuracy of the four corner regions of each area is used to analyze the influence pattern of neighborhood information.

(c-e) The images corresponding to the randomly selected red detection regions are obtained by cropping from a large-extent image allows obtaining. These red regions exhibit the same land features but different neighborhood information.

### 3.1.3. Scale information analysis dataset

To analyze the influence of scale information on the effectiveness of open-pit mine CD in HR remote sensing images, a scale information analysis dataset was created by modifying the test set of open-pit mine CD dataset. The test set was sequentially downsampled by factors of 2, 4, 8, and 16, and then restored to the standard image size, thereby reducing image resolution with maintaining the pixel count. Combined with the original scale, a total of five scale-specific test sets were generated to analyze the influence pattern of scale information.

## 3.2. Experimental settings

### 3.2.1. Implementation details

The experiments were implemented using a desktop computer with an 11th Gen Intel i7-11700, 64GB memory, and NVIDIA GeForce RTX3060. The deep learning framework adopted was Pytorch, written in python. During CD models training, the batch size was set to eight, the initial learning rate was 0.001, the optimizer was Adam, and 200 epochs were trained.

### 3.2.2. Evaluation metrics

The purpose of CD is to determine the changed pixels and the unchanged pixels. In essence, it can be classified as a binary classification problem. To quantitatively evaluate the effectiveness of the data augmentation methods quantitatively, four evaluation metrics were calculated: recall (Rec), precision (Pre), overall accuracy (OA), intersection over union (IoU), and F1-score (F1). Higher values of these metrics indicate better detection results.

$$Rec = \frac{TP}{TP + FN} \quad (2)$$

$$Pre = \frac{TP}{TP + FP} \quad (3)$$

$$OA = \frac{TP + TN}{TP + TN + FP + FN} \quad (4)$$

$$IoU = \frac{TP}{TP + FP + FN} \quad (5)$$

$$F1 = \frac{2}{Rec^{-1} + Pre^{-1}} \quad (6)$$

where TP, TN, FP, and FN represent the numbers of true positives, true negatives, false positives, and false negatives, respectively.

Notably, the F1 and IoU simultaneously considers precision and recall, rendering it more suitable for detecting imbalanced samples in the foreground and background and thus serving as the more reliable evaluation metric.

### 3.2.3. Comparative methods

To verify the superiority of INSINet, several state-of-the-art (SOTA) methods for remote sensing image CD are adopted as comparative methods, including FC-EF (Caye Daudt et al., 2018), FC-Siam-Di (Caye Daudt et al., 2018), FC-Siam-Conc(Caye Daudt et al., 2018), SNUNet (Fang et al., 2022), BIT (H. Chen et al., 2022), TinyCD (Codegoni et al., 2023), DMINet (Feng et al., 2023), USSFCNet (Lei et al., 2023). The following is a brief introduction to each method.

(1) FC-EF: a fully convolutional early fusion U-shape network.

(2) FC-Siam-Di: a fully convolutional Siamese-difference network, extending FC-EF.

(3) FC-Siam-Conc: a fully convolutional Siamese-concatenation network, extending FC-EF.

(4) SNUNet: a densely connected Siamese nested U-shape network based on U-Net++ (Zhou et al., 2018).

(5) BIT: a bitemporal image CD network based on Transformer (Vaswani et al., 2017).

(6) TinyCD: a lightweight and effective Siamese U-shape CD network.

(7) DMINet: a dual-branch multilevel intertemporal Siamese network.

(8) USSFCNet: an efficient ultralightweight spatial–spectral feature cooperation Siamese U-shape network.

### 3.3. Neighborhood and scale information analysis

In the analysis of the influence of neighborhood and scale information on the open-pit mine CD in HR remote sensing images, the mean of F1 values of each existing CD model was used as a comprehensive analysis result. Considering the similarity among the FC-EF, FC-Siam-Di and FC-Siam-Conc models, only FC-Siam-Conc was selected for the analysis. In addition, the effectiveness of neighborhood and scale information for INSINet was also demonstrated.

#### 3.3.1. Neighborhood information analysis

The influence pattern of neighborhood information on CD models was analyzed using the neighborhood information analysis dataset, as shown in Fig. 7(a). Different neighborhood information resulted in varying CD accuracies for the same detection region. When the detection region was located in the middle ring, inner ring, and core area, the CD accuracy remained relatively stable. However, a significant decrease in CD accuracy was observed when the detection area was located in the outer ring. This suggests that the absence of neighborhood information impacts the effectiveness of CD in open-pit areas, with the main effect concentrated in the outer ring.

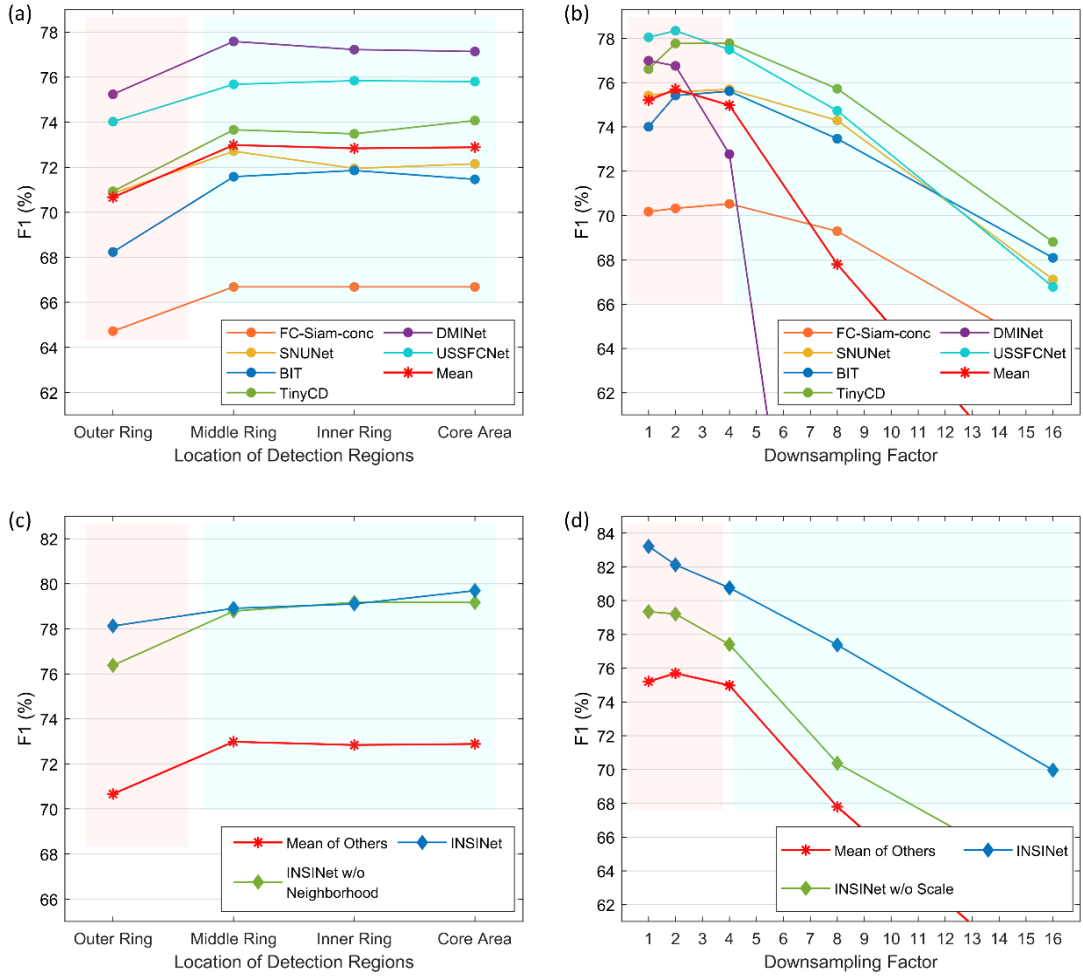

Fig. 7. Analysis of neighborhood and scale information on the effectiveness of open-pit mine CD. Red areas highlight challenges in effectively utilizing neighborhood and scale information, while blue areas remain largely unaffected. (a) Neighborhood information analysis. (b) Scale information analysis. (c) Effectiveness of neighborhood information on INSINet. (d)Effectiveness of scale information on INSINet.

### 3.3.2. Scale information analysis

The influence pattern of scale information on CD models was analyzed using the scale information analysis dataset, as shown in Fig. 7(b). Unlike general CD detection, in HR open-pit mine CD, higher resolution did not necessarily result in better detection results. As the resolution decreases, the CD accuracy generally showed a trend of slight improvement followed by rapid decline. This suggested that in most cases, the optimal scale for HR open pit mine CD might not be 1 m, but 2 m or more. This could be related to the size of the detection targets in the mining area, as detailed in section 4.2. In addition, the selection of the optimal scale is also related to the structure of the CD model, and various factors jointly determine the selection of the optimal scale. In fact, a more effective approach is to further deepen the acquisition and fusion process of multi-scale information to realize intelligent selection and utilization of scale information.

### 3.3.3. Effectiveness evaluation of neighborhood and scale information for INSINet

To evaluate the effectiveness of neighborhood and scale information on INSINet, ablation experiments were conducted by removing corresponding components in INSINet. As shown in Fig. 7(c), without utilizing neighborhood information, the performance of INSINet was influenced similarly by the amount of neighborhood information as other methods. However, with the incorporation of neighborhood information, INSINet showed improved detection performance in the outer ring, with further enhancement in the core area. This indicates that the neighborhood information can not only enhance detection at the image boundary region but also provide a larger receptive field beneficial for detection in other regions of the image. As depicted in Fig. 7(d), without utilizing scale information, the accuracy of INSINet was similar to that of other methods. In this scenario, the optimal scale for INSINet was 1 or 2 m. However, after integrating multi-scale information, the overall performance of INSINet was greatly enhanced, achieving better detection results. At this case, the performance of INSINet correlated positively with increasing resolution. It shows that INSINet effectively combines the optimal scale information and HR image information, further improving the model effectiveness. More discussion detailed in Section 4.2.

### 3.4. Method comparison

Quantitative comparison results between INSINet and existing SOTA methods on the open-pit mine CD dataset are presented in Table 1. INSINet achieved the best performance with 97.69% for OA, 71.26% for IoU and 83.22% for F1, surpassing the second-ranked method by 0.47%, 7.18%, and 5.17%, respectively. The reason for the superior performance of INSINet could be understood by analyzing Pre and Rec. Other methods exhibited a significant imbalance between Pre and Rec, with Pre typically outperforming Rec by 10-30 percentage points. This was attributed to the imbalanced nature of foreground-background in open-pit mine CD tasks, where achieving high Rec is more challenging than Pre, prompting other methods to adopt conservative detection strategies (Xie et al., 2024). In contrast, INSINet exceled in both Pre and Rec, demonstrating a more accurate understanding of mining areas and fewer detection omissions. In practical applications of open-pit mine CD in HR remote sensing images, since the mining areas occupies a small portion and the change areas even smaller, Rec is often more important than Pre. Therefore, INSINet holds greater practical value compared to other methods.

TABLE I

Quantitative comparison results of different methods.

| Method | Pre (%) | Rec (%) | OA (%) | IoU (%) | F1 (%) |
| --- | --- | --- | --- | --- | --- |
| FCEF | 84.17 | 61.77 | 96.55 | 54.81 | 70.71 |
| FC-Siam-Diff | 86.59 | 55.59 | 96.35 | 50.70 | 67.02 |
| FC-Siam-Conc | 85.24 | 60.10 | 96.55 | 54.17 | 70.18 |
| SNUNet | 80.61 | 72.10 | 96.77 | 60.83 | 75.42 |
| BIT | 81.05 | 68.87 | 96.73 | 58.88 | 74.01 |
| TinyCD | 84.42 | 70.79 | 97.09 | 62.24 | 76.61 |
| DMINet | **86.62** | 69.34 | 97.18 | 62.60 | 76.99 |
| USSFCNet | 84.62 | 72.82 | 97.22 | 64.08 | 78.05 |
| INSINet | 82.48 | **84.17** | **97.69** | **71.26** | **83.22** |

The qualitative comparison between INSINet and existing SOTA methods is illustrated in Fig. 8. It could be observed that INSINet exhibited significant advantages in Rec while maintaining Pre. By leveraging neighborhood information, INSINet demonstrated superior capability in detecting changes both at the image center and boundary regions, ensuring consistency in detection results at the stitching seams. Moreover, through exploitation of multi-scale information, INSINet exhibited favorable performance across targets of various sizes, resulting in higher integrity and reduced noise in the detection results, as detailed in Section 4.2.

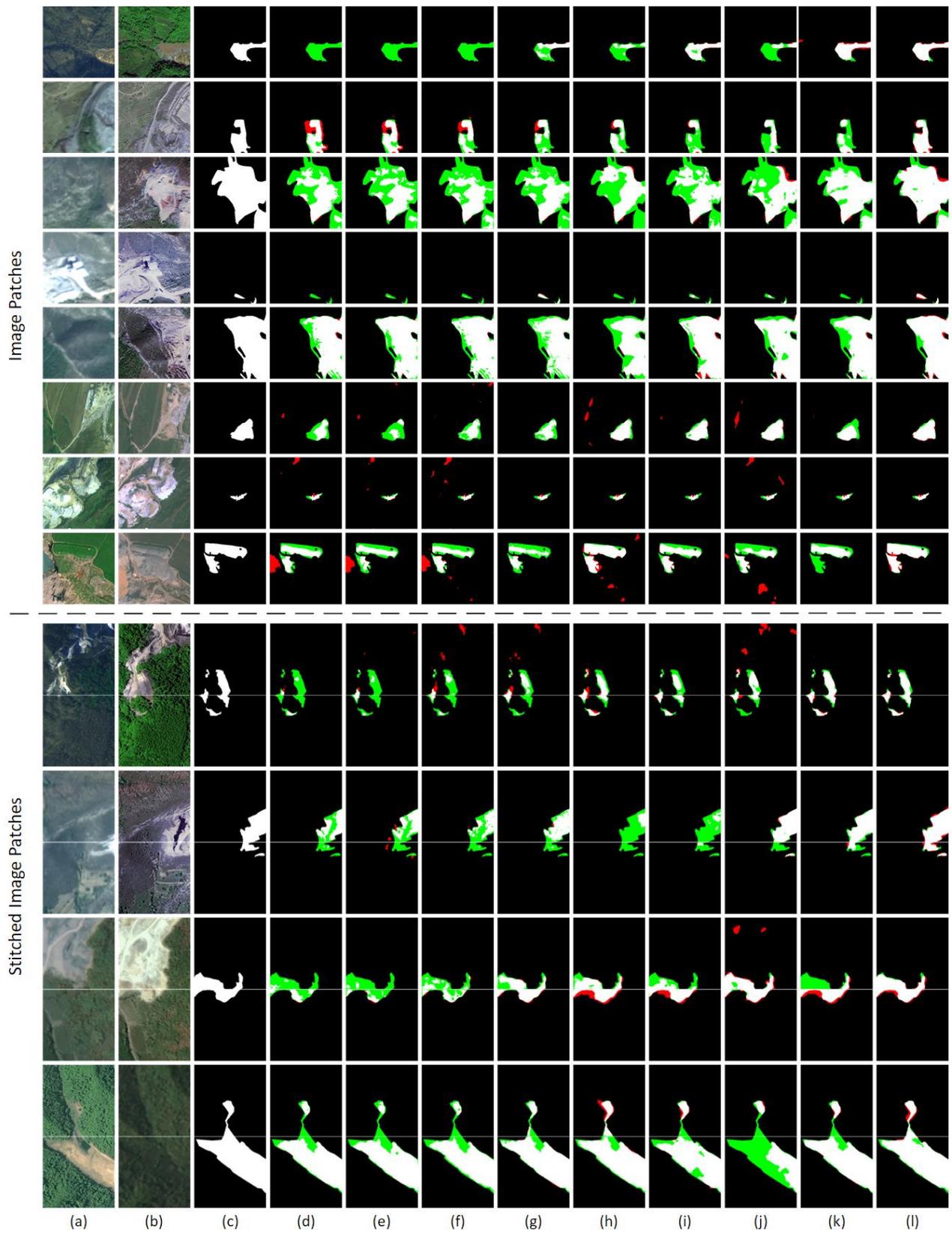

Fig. 8. Qualitative comparison results of different methods. Different colors are used for a better view, i.e., white for TP, black for TN, red for FP, and green for FN. (a) Image at T1. (b) Image at T2. (c) Ground

truth. (d) FCEF. (e) FC-Siam-Diff. (f) FC-Siam-Conc. (g) SNUNet. (h) BIT. (i) TinyCD. (j) DMINet. (k) USSFCNet. (l) INSINet.

### 3.5. Model efficiency analysis

The computational efficiency between the proposed INSINet and the existing SOTA methods were tested and analyzed. Three metrics were used for the analysis, including the model effectiveness F1, parameter count (Params) and time consumption (MACs). The analysis results are presented in Table 2. INSINet achieved the best model effectiveness and also obtain sub-optimal and optimal results in terms of Params and MACs, respectively. Overall, INSINet demonstrated the best performance in model efficiency.

TABLE II

Efficiency analysis results of different methods.

| Method | F1 (%) | Params (M) | MACs (G) |
| --- | --- | --- | --- |
| FCEF | 70.71 | 1.35 | 3.55 |
| FC-Siam-Diff | 67.02 | 1.35 | 4.69 |
| FC-Siam-Conc | 70.18 | 1.99 | 5.89 |
| SNUNet | 75.42 | 12.03 | 54.84 |
| BIT | 74.01 | 3.50 | 10.63 |
| TinyCD | 76.61 | **0.29** | 1.54 |
| DMINet | 76.99 | 6.24 | 14.55 |
| USSFCNet | 78.05 | 1.52 | 4.86 |
| INSINet | **83.22** | 0.83 | **1.32** |

### 3.6. Ablation study

Components of INSINet were subjected to ablation experiments to demonstrate the rationality and effectiveness of the proposed method. Ablation experiments were also employed to analyze the effectiveness of neighborhood and scale information on INSINet. Analysis metrics included the F1, Params and MACs. The results of this analysis are presented in Table 3. It was evident that each component of INSINet contributed to its detection effectiveness. However, except for CNF, most components led to an increase in the number of parameters and time consumption of INSINet. Combining Table 2 and 3, although the addition of some components made INSINet less lightweight, it still outperformed existing models in terms of execution efficiency, making it acceptable. In addition, neighborhood and scale information contributed to an increase in F1 values of 3.08% and 3.32% for INSINet, respectively, resulting in a total improvement of 6.40%. This demonstrated the significance of both factors.

TABLE III

Ablation Study of INSINet.

| Baseline (Siamese Network) | TF | Scale Information | | Neighborhood Information | | F1 (%) | Params (M) | MACs (G) |
|---|---|---|---|---|---|---|---|---|
| | | MDSA | | Quadruplet Network | CNF | | | |
| √ | | | | | | 75.45 | 0.27 | 0.48 |
| √ | √ | | | | | 76.82 | 0.36 | 0.62 |
| √ | √ | √ | | | | 80.14 | 0.50 | 0.75 |
| √ | √ | √ | | √ | | 82.15 | 0.90 | 1.31 |
| √ | √ | √ | | √ | √ | 83.22 | 0.83 | 1.32 |

## 4. Discussion

### 4.1. Effectiveness evaluation of neighborhood information on CD of unregistered images

Despite the pre-registration of CD, there may still be instances of incomplete registration in practical application scenarios. Neighborhood information provides a broader perspective, potentially aiding in the detection of unregistered images. To further analyze the effect of neighborhood information on the detection of unregistered images, experiments were conducted using original unregistered images on the basis of the previous neighborhood information analysis experiments, and results were compared with those shown in Fig. 7. Fig. 9 (a) illustrates the experimental results, while Fig. 9 (b) quantifies the adverse effect of unregistered images on CD. It was observed that without the utilization of neighborhood information, the performance of both INSINet and other methods was influenced by the accuracy of image registration. However, after incorporating neighborhood information, the performance of INSINet was largely unaffected by registration accuracy. This indicates that neighborhood information mitigates the interference of image mismatch in CD by enlarging the receptive field.

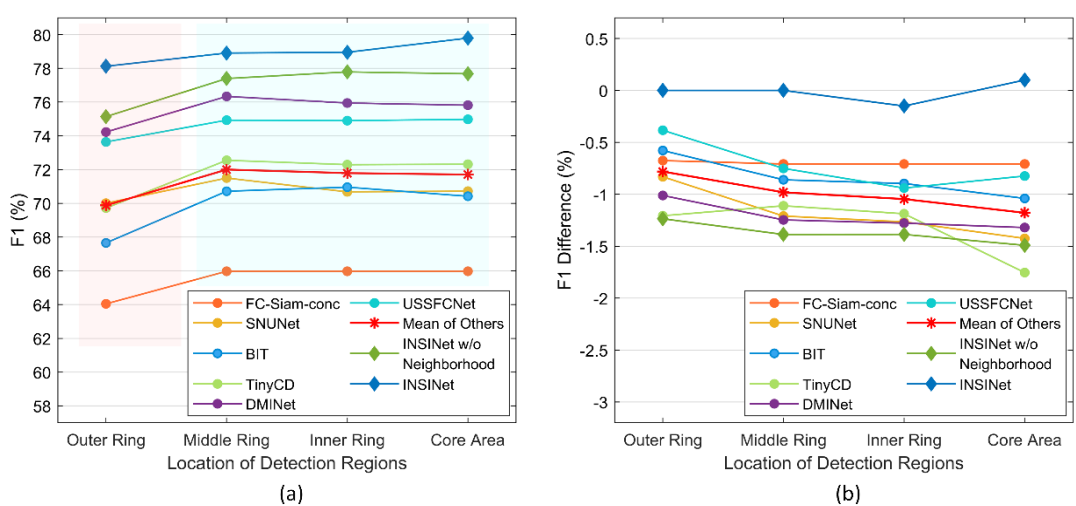

Fig. 9. Analysis of the impact of neighborhood information on the CD of unregistered images. (a) Accuracy of CD for unregistered images. Red areas highlight challenges in effectively utilizing

neighborhood information, while blue areas remain largely unaffected. (b) Difference in CD accuracy between unregistered and registered images.

## 4.2. Effectiveness evaluation of scale information on CD of different sized target

To further analyze the application effect of scale information, on the basis of the previous scale information analysis experiments, the test set was further divided into three subsets according to the size of the detection target, and the results are shown in Fig. 10. It could be seen that for small targets, the higher the image resolution, the better the detection effect. For medium targets, under detecting HR images, an increase in resolution did not bring a more significant improvement in CD performance. For large targets, the increase in resolution might instead lead to a decrease in CD performance. This suggests that the optimal scales for different sizes of targets in open-pit mine CD varies, and reducing resolution might help with the detection of larger targets. Therefore, the integration of multi-scale information is beneficial to augment CD effectiveness. In addition, when INSINet operated without utilizing scale information, its performance was consistent with other methods. However, after using scale information, the performance of INSINet improved with increasing resolution for targets of any size. This indicates that INSINet achieves comprehensive exploitation and fusion of multi-scale information, thus mitigating the influence of target size.

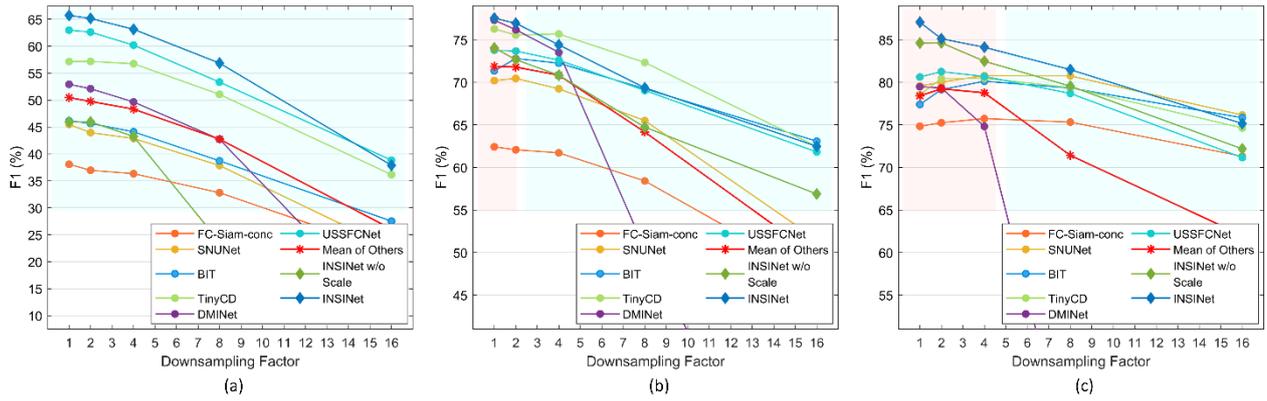

Fig. 10. Analysis of the impact of scale information on the CD for targets of different sizes. Red areas highlight challenges in effectively utilizing scale information, while blue areas remain largely unaffected. (a) Accuracy of CD for small targets. (b) Accuracy of CD for medium targets. (c) Accuracy of CD for large targets.

Additionally, as shown in Fig. 10, regardless of the utilization of scale information, INSINet achieved significantly better CD accuracy for large targets compared to small and medium targets. This improvement could be attributed to the incorporation of neighborhood information. Neighborhood information allows INSINet to obtain a larger receptive field, enhancing its ability to detect extensive changes. However, the incorporation of neighborhood information also leads to a tendency for INSINet without scale information tend to focus more on large targets over small ones. Therefore, it is necessary to integrate both neighborhood and scale information for optimal performance.

## 5. Conclusion

Considering the influence of neighborhood and scale factors on open-pit CD in HR remote sensing images, this study proposed INSINet for CD based on an exploration of the influence patterns of neighborhood and scale information. Specifically, INSINet acquires neighborhood information from the 8-neighborhood images around the target image, which is then input into the Quadruple network to assist in feature extraction of the central image. Subsequently, time features fusion and center-neighborhood features fusion are achieved by TF and CNF modules, respectively. Finally, the MDSA module facilitates multi-scale feature aggregation to detect the changes. Experimental validation on both neighborhood information and scale information analysis datasets demonstrates the efficacy of INSINet, with a 6.40% increase in F1 score attributed to the aggregation of these factors. Quantitative and qualitative analyses on the open-pit mine CD dataset confirmed the reliability of INSINet. INSINet achieved SOTA performance with OA, IoU, and F1 of 97.69%, 71.26%, and 83.22%, respectively. In addition, INSINet also considered the model lightweight and demonstrated optimal performance in execution efficiency. The discussion section illustrated the effects of neighborhood and scale information on unregistered images and targets of different sizes, further affirming the effectiveness of INSINet.

## References

Barenblitt, A., Payton, A., Lagomasino, D., Fatoyinbo, L., Asare, K., Aidoo, K., Pigott, H., Som, C.K., Smeets, L., Seidu, O., Wood, D., 2021. The large footprint of small-scale artisanal gold mining in Ghana. Science of The Total Environment 781, 146644. https://doi.org/10.1016/j.scitotenv.2021.146644

Blachowski, J., Dynowski, A., Buczyńska, A., Ellefmo, S.L., Walerysiak, N., 2023. Integrated Spatiotemporal Analysis of Vegetation Condition in a Complex Post-Mining Area: Lignite Mine Case Study. Remote Sensing 15, 3067. https://doi.org/10.3390/rs15123067

Camalan, S., Cui, K., Pauca, V.P., Alqahtani, S., Silman, M., Chan, R., Plemmons, R.J., Dethier, E.N., Fernandez, L.E., Lutz, D.A., 2022. Change Detection of Amazonian Alluvial Gold Mining Using Deep Learning and Sentinel-2 Imagery. Remote Sensing 14, 1746. https://doi.org/10.3390/rs14071746

Caye Daudt, R., Le Saux, B., Boulch, A., 2018. Fully Convolutional Siamese Networks for Change Detection, in: 2018 25th IEEE International Conference on Image Processing (ICIP). Presented at the 2018 25th IEEE International Conference on Image Processing (ICIP), IEEE, Athens, Greece, pp. 4063–4067. https://doi.org/10.1109/ICIP.2018.8451652

Chen, H., Qi, Z., Shi, Z., 2022. Remote Sensing Image Change Detection With Transformers. IEEE Transactions on Geoscience and Remote Sensing 60, 1–14. https://doi.org/10.1109/TGRS.2021.3095166

Chen, M., Zhang, Q., Ge, X., Xu, B., Hu, H., Zhu, Q., Zhang, X., 2023. A Full-Scale Connected CNN–Transformer Network for Remote Sensing Image Change Detection. Remote Sensing 15, 5383. https://doi.org/10.3390/rs15225383

Chen, T., Lu, Z., Yang, Y., Zhang, Y., Du, B., Plaza, A., 2022. A Siamese Network Based U-Net for Change Detection in High Resolution Remote Sensing Images. IEEE Journal of Selected Topics in Applied Earth Observations and Remote Sensing 15, 2357–2369. https://doi.org/10.1109/JSTARS.2022.3157648

Chen, W.T., Li, X.J., He, H.X., Wang, L.Z., 2018. A Review of Fine-Scale Land Use and Land Cover Classification in Open-Pit Mining Areas by Remote Sensing Techniques. Remote Sensing 10, 15. https://doi.org/10.3390/rs10010015

Chen, Y., Fan, K., Chang, Y., Moriyama, T., 2023. Special Issue Review: Artificial Intelligence and Machine Learning Applications in Remote Sensing. Remote Sensing 15, 569. https://doi.org/10.3390/rs15030569

Codegoni, A., Lombardi, G., Ferrari, A., 2023. TINYCD: a (not so) deep learning model for change detection. Neural Comput & Applic 35, 8471–8486. https://doi.org/10.1007/s00521-022-08122-3

Deng, Y., Meng, Y., Chen, Jingbo, Yue, A., Liu, D., Chen, Jing, 2023. TChange: A Hybrid Transformer-CNN Change Detection Network. Remote Sensing 15, 1219. https://doi.org/10.3390/rs15051219

Ding, L., Zhang, J., Bruzzone, L., 2020. Semantic Segmentation of Large-Size VHR Remote Sensing Images Using a Two-Stage Multiscale Training Architecture. IEEE Trans. Geosci. Remote Sensing 58, 5367–5376. https://doi.org/10.1109/TGRS.2020.2964675

Du, Shouhang, Li, W., Li, J., Du, Shihong, Zhang, C., Sun, Y., 2022a. Open-pit mine change detection from high resolution remote sensing images using DA-UNet++ and object-based approach. International Journal of Mining, Reclamation and Environment 36, 512–535. https://doi.org/10.1080/17480930.2022.2072102

Du, Shouhang, Xing, J., Li, J., Du, Shihong, Zhang, C., Sun, Y., 2022b. Open-Pit Mine Extraction from Very High-Resolution Remote Sensing Images Using OM-DeepLab. Natural Resources Research 31, 3173–3194. https://doi.org/10.1007/s11053-022-10114-y

Du, Shouji, Du, Shihong, Liu, B., Zhang, X., 2021. Incorporating DeepLabv3+ and object-based image analysis for semantic segmentation of very high resolution remote sensing images. International Journal of Digital Earth 14, 357–378. https://doi.org/10.1080/17538947.2020.1831087

Everingham, M., Eslami, S.M.A., Van Gool, L., Williams, C.K.I., Winn, J., Zisserman, A., 2015. The Pascal Visual Object Classes Challenge: A Retrospective. Int J Comput Vis 111, 98–136. https://doi.org/10.1007/s11263-014-0733-5

Fang, S., Li, K., Shao, J., Li, Z., 2022. SNUNet-CD: A Densely Connected Siamese Network for Change Detection of VHR Images. IEEE Geosci. Remote Sensing Lett. 19, 1–5. https://doi.org/10.1109/LGRS.2021.3056416

Fekete, A., Cserep, M., 2021. Tree segmentation and change detection of large urban areas based on airborne LiDAR. Computers & Geosciences 156, 104900. https://doi.org/10.1016/j.cageo.2021.104900

Feng, Y., Jiang, J., Xu, H., Zheng, J., 2023. Change Detection on Remote Sensing Images Using Dual-Branch Multilevel Intertemporal Network. IEEE Transactions on Geoscience and Remote Sensing 61, 1–15. https://doi.org/10.1109/TGRS.2023.3241257

Gallwey, J., Robiati, C., Coggan, J., Vogt, D., Eyre, M., 2020. A Sentinel-2 based multispectral convolutional neural network for detecting artisanal small-scale mining in Ghana: Applying deep learning to shallow mining. Remote Sensing of Environment 248, 111970. https://doi.org/10.1016/j.rse.2020.111970

Ganaie, M.A., Hu, M., Malik, A.K., Tanveer, M., Suganthan, P.N., 2022. Ensemble deep learning: A review. Engineering Applications of Artificial Intelligence 115, 105151. https://doi.org/10.1016/j.engappai.2022.105151

Gao, S., Chen, Y., Li, K., Li, Y., Yu, J., Rao, R., 2021. Mapping Opencast Iron Mine and Mine Solid Waste Based on a New Spectral Index From Medium Spatial Resolution Satellite Data. IEEE Journal of Selected Topics in Applied Earth Observations and Remote Sensing 14, 7788–7798. https://doi.org/10.1109/JSTARS.2021.3098801

Hao, F., Ma, Z.-F., Tian, H.-P., Wang, H., Wu, D., 2023. Semi-supervised label propagation for multi-source remote sensing image change detection. Computers & Geosciences 170, 105249. https://doi.org/10.1016/j.cageo.2022.105249

Howard, A., Sandler, M., Chen, B., Wang, W., Chen, L.-C., Tan, M., Chu, G., Vasudevan, V., Zhu, Y., Pang, R., Adam, H., Le, Q., 2019. Searching for MobileNetV3, in: 2019 IEEE/CVF International Conference on Computer Vision (ICCV). Presented at the 2019 IEEE/CVF International Conference on Computer Vision (ICCV), pp. 1314–1324. https://doi.org/10.1109/ICCV.2019.00140

Jiang, H., Peng, M., Zhong, Y., Xie, H., Hao, Z., Lin, J., Ma, X., Hu, X., 2022. A Survey on Deep Learning-Based Change Detection from High-Resolution Remote Sensing Images. Remote Sensing 14, 1552. https://doi.org/10.3390/rs14071552

Kumar, A., Gorai, A.K., 2023. Design of an optimized deep learning algorithm for automatic classification of high-resolution satellite dataset (LISS IV) for studying land-use patterns in a mining region. Computers & Geosciences 170, 105251. https://doi.org/10.1016/j.cageo.2022.105251

Lei, T., Geng, X., Ning, H., Lv, Z., Gong, M., Jin, Y., Nandi, A.K., 2023. Ultralightweight Spatial–Spectral Feature Cooperation Network for Change Detection in Remote Sensing Images. IEEE Transactions on Geoscience and Remote Sensing 61, 1–14. https://doi.org/10.1109/TGRS.2023.3261273

Li, X., Yan, L., Zhang, Y., Mo, N., 2022. SDMNet: A Deep-Supervised Dual Discriminative Metric Network for Change Detection in High-Resolution Remote Sensing Images. IEEE Geosci. Remote Sensing Lett. 19, 1–5. https://doi.org/10.1109/LGRS.2022.3216627

Liu, S., Ding, W., Liu, C., Liu, Y., Wang, Y., Li, H., 2018. ERN: Edge Loss Reinforced Semantic Segmentation Network for Remote Sensing Images. Remote. Sens. 10, 1339.

Liu, Y., Wang, X., Zhang, Z., Deng, F., 2023. Deep learning in image segmentation for mineral production: A review. Computers & Geosciences 180, 105455. https://doi.org/10.1016/j.cageo.2023.105455

Liu, Y., Zhang, Z., Liu, X., Wang, L., Xia, X., 2021. Deep learning-based image classification for online multi-coal and multi-class sorting. Computers & Geosciences 157, 104922. https://doi.org/10.1016/j.cageo.2021.104922

Lv, Z., Liu, T., Benediktsson, J.A., Falco, N., 2022. Land Cover Change Detection Techniques: Very-high-resolution optical images: A review. IEEE Geosci. Remote Sens. Mag. 10, 44–63. https://doi.org/10.1109/MGRS.2021.3088865

Nie, X., Hu, Z., Ruan, M., Zhu, Q., Sun, H., 2022. Remote-Sensing Evaluation and Temporal and Spatial Change Detection of Ecological Environment Quality in Coal-Mining Areas. Remote Sensing 14, 345. https://doi.org/10.3390/rs14020345

Padmanaban, R., Bhowmik, A.K., Cabral, P., 2017. A Remote Sensing Approach to Environmental Monitoring in a Reclaimed Mine Area. IJGI 6, 401. https://doi.org/10.3390/ijgi6120401

Pan, X., Zhang, P., Guo, S., Zhang, W., Xia, Z., Fang, H., Du, P., 2023. A Novel Exposed Coal Index Combining Flat Spectral Shape and Low Reflectance. IEEE Trans. Geosci. Remote Sensing 61, 1–16. https://doi.org/10.1109/TGRS.2023.3333568

Romary, T., Ors, F., Rivoirard, J., Deraisme, J., 2015. Unsupervised classification of multivariate geostatistical data: Two algorithms. Computers & Geosciences 85, 96–103. https://doi.org/10.1016/j.cageo.2015.05.019

Ronneberger, O., Fischer, P., Brox, T., 2015. U-Net: Convolutional Networks for Biomedical Image Segmentation. https://doi.org/10.48550/arXiv.1505.04597

S. Ji, S. Wei, M. Lu, 2019. Fully Convolutional Networks for Multisource Building Extraction From an Open Aerial and Satellite Imagery Data Set. ITGRS 57, 574–586. https://doi.org/10.1109/TGRS.2018.2858817

Song, K., Jiang, J., 2021. AGCDetNet:An Attention-Guided Network for Building Change Detection in High-Resolution Remote Sensing Images. IEEE Journal of Selected Topics in Applied Earth Observations and Remote Sensing 14, 4816–4831. https://doi.org/10.1109/JSTARS.2021.3077545

Song, R., Ni, W., Cheng, W., Wang, X., 2022. CSANet: Cross-Temporal Interaction Symmetric Attention Network for Hyperspectral Image Change Detection. IEEE Geoscience and Remote Sensing Letters 19, 1–5. https://doi.org/10.1109/LGRS.2022.3179134

Specification for Investigation and Evaluation of Mine Geological Environment, 2014.

Tang, C., Zhang, Z., He, G., Long, T., Wang, G., Wei, M., She, W., 2021. An improved fully convolution network model for change detection in mining areas using sentinel-2 images. Remote Sensing Letters 12, 684–694. https://doi.org/10.1080/2150704X.2021.1925372

Technical Requirements for Remote Sensing Monitoring of Mineral Resource Development, 2011.

Varghese, A., Gubbi, J., Ramaswamy, A., Balamuralidhar, P., 2019. ChangeNet: A Deep Learning Architecture for Visual Change Detection, in: Leal-Taixé, L., Roth, S. (Eds.), Computer Vision – ECCV 2018 Workshops. Springer International Publishing, Cham, pp. 129–145. https://doi.org/10.1007/978-3-030-11012-3_10

Vaswani, A., Shazeer, N.M., Parmar, N., Uszkoreit, J., Jones, L., Gomez, A.N., Kaiser, L., Polosukhin, I., 2017. Attention is All you Need, in: Neural Information Processing Systems.

Wang, D., Zhang, C., Han, M., 2022. MLFC-net: A multi-level feature combination attention model for remote sensing scene classification. Computers & Geosciences 160, 105042. https://doi.org/10.1016/j.cageo.2022.105042

Wang, L., Lee, C., Tu, Z., Lazebnik, S., 2015. Training Deeper Convolutional Networks with Deep Supervision.

Wang, Z., Wang, J., Yang, K., Wang, L., Su, F., Chen, X., 2022. Semantic segmentation of high-resolution remote sensing images based on a class feature attention mechanism fused with Deeplabv3+. Computers & Geosciences 158, 104969. https://doi.org/10.1016/j.cageo.2021.104969

Wilkins, A.H., Strange, A., Duan, Y., Luo, X., 2020. Identifying microseismic events in a mining scenario using a convolutional neural network. Computers & Geosciences 137, 104418. https://doi.org/10.1016/j.cageo.2020.104418

Woo, S., Park, J., Lee, J.-Y., Kweon, I.S., 2018. CBAM: Convolutional Block Attention Module, in: Ferrari, V., Hebert, M., Sminchisescu, C., Weiss, Y. (Eds.), Computer Vision – ECCV 2018, Lecture Notes in Computer Science. Springer International Publishing, Cham, pp. 3–19. https://doi.org/10.1007/978-3-030-01234-2_1

Wu, X., Yang, L., Ma, Y., Wu, C., Guo, C., Yan, H., Qiao, Z., Yao, S., Fan, Y., 2023. An end-to-end multiple side-outputs fusion deep supervision network based remote sensing image change detection algorithm. Signal Processing 213, 109203. https://doi.org/10.1016/j.sigpro.2023.109203

Xiang, X., Tian, D., Lv, N., Yan, Q., 2022. FCDNet: A Change Detection Network Based on Full-Scale Skip Connections and Coordinate Attention. IEEE Geoscience and Remote Sensing Letters 19, 1–5. https://doi.org/10.1109/LGRS.2022.3184179

Xie, Z., Jiang, J., Yuan, D., Li, K., Liu, Z., 2024. GAN-Based Sub-Instance Augmentation for Open-Pit Mine Change Detection in Remote Sensing Images. IEEE Transactions on Geoscience and Remote Sensing 62, 1–19. https://doi.org/10.1109/TGRS.2023.3336658

Xie, Z., Jiang, J., Yuan, D., Li, K., Liu, Z., 2023. GAN-based Sub-instance Augmentation for Open-pit Mine Change Detection in Remote Sensing Images. IEEE Transactions on Geoscience and Remote Sensing 1–1. https://doi.org/10.1109/TGRS.2023.3336658

Zhang, C., Yue, P., Tapete, D., Jiang, L., Shangguan, B., Huang, L., Liu, G., 2020. A deeply supervised image fusion network for change detection in high resolution bi-temporal remote sensing images. ISPRS Journal of Photogrammetry and Remote Sensing 166, 183–200. https://doi.org/10.1016/j.isprsjprs.2020.06.003

Zhang, C., Zuo, R., Xiong, Y., Zhao, X., Zhao, K., 2022. A geologically-constrained deep learning algorithm for recognizing geochemical anomalies. Computers & Geosciences 162, 105100. https://doi.org/10.1016/j.cageo.2022.105100

Zhang, Z., Lu, W., Cao, J., Xie, G., 2022. MKANet: An Efficient Network with Sobel Boundary Loss for Land-Cover Classification of Satellite Remote Sensing Imagery. Remote Sensing 14, 4514. https://doi.org/10.3390/rs14184514

Zhou, Z., Rahman Siddiquee, M.M., Tajbakhsh, N., Liang, J., 2018. UNet++: A Nested U-Net Architecture for Medical Image Segmentation, in: Deep Learning in Medical Image Analysis and Multimodal Learning for Clinical Decision Support. Presented at the 4th International

Workshop on Deep Learning in Medical Image Analysis (DLMIA) / 8th International Workshop on Multimodal Learning for Clinical Decision Support (ML-CDS), Springer International Publishing, Cham, pp. 3–11. https://doi.org/10.1007/978-3-030-00889-5_1